%% file: emnlp2021.tex
\pdfoutput=1

\documentclass[11pt]{article}

\usepackage[]{emnlp2021}

\usepackage{times}
\usepackage{latexsym}
\usepackage{amssymb}
\usepackage{amsmath}
\usepackage{amsfonts}
\usepackage{todonotes}
\usepackage{mathtools}
\usepackage{multirow} 
\usepackage[ruled,vlined,linesnumbered]{algorithm2e}

\usepackage[T1]{fontenc}

\usepackage[utf8]{inputenc}

\usepackage{microtype}

\let\norm\undefined 
\DeclarePairedDelimiter\norm{\lVert}{\rVert}

\title{Don't Search for a Search Method --- \\Simple Heuristics Suffice for Adversarial Text Attacks}
\author{Nathaniel Berger$^\dagger$\Thanks{ Part of the work done while the first author was interning at Google.} \and Stefan Riezler$^{\dagger,\ddagger}$ \\
  $^\dagger$Computational Linguistics \& $^\ddagger$IWR \\
  Heidelberg University, Germany \\
  \And
  \mbox{~~~~~~~~~~~~~~}Artem Sokolov$^{\diamondsuit,\dagger}$ \and Sebastian Ebert$^\diamondsuit$ \\
  $^\diamondsuit$Google Research\\Berlin, Germany
  \AND
  {\normalfont \texttt{\{berger,riezler\}@cl.uni-heidelberg.de}, \texttt{\{artemsok,eberts\}@google.com}} }

\begin{document}
\maketitle
\begin{abstract}
Recently more attention has been given to adversarial attacks on neural networks for natural language processing (NLP). A central research topic has been the investigation of search algorithms and search constraints, accompanied by benchmark algorithms and tasks.
We implement an algorithm inspired by zeroth order optimization-based attacks and compare with the benchmark results in the TextAttack framework. Surprisingly, we find that optimization-based methods do not yield any improvement in a constrained setup and slightly benefit from approximate gradient information only in unconstrained setups where search spaces are larger. In contrast, simple heuristics exploiting nearest neighbors without querying the target function yield substantial success rates in constrained setups, and nearly full success rate in unconstrained setups, at an order of magnitude fewer queries.
We conclude from these results that current TextAttack benchmark tasks are too easy and constraints are too strict, preventing meaningful research on black-box adversarial text attacks.
\end{abstract}

\section{Introduction}

Neural networks are currently deployed in many production systems from image classification to natural language processing (NLP). While they show impressive results, many of such systems are also susceptible to adversarial examples. These examples are close in space to data points that are classified correctly, but due to small changes are classified incorrectly. This corresponds to the local non-smoothness noted by \citet{szegedy2014intriguing}. 

 \citet{chen2017zoo} present a zeroth order optimization-based (ZOO) attack that significantly outperforms other black-box attacks for image processing. An initial goal of our work was to investigate whether a ZOO approach that guides the search for adversarial examples by approximate gradient information can be transferred to adversarial attacks in NLP. To this end,  we implemented a ZOO-inspired algorithm in the TextAttack framework \citep{morris2020textattack} and compared it with their benchmark results \citep{yoo2020searching}. Surprisingly, we found that search guided by a ZOO approach only yields minimal improvements, whereas heuristic search exploiting nearest neighbors yields competitive success rates at minimal query count. 
We conclude from these results that the victim models in current benchmark tasks are too easily broken and imperceptibility constraints are too strict, thus defeating meaningful research on black-box search methods for adversarial attacks.

\section{Related Work}

Black-box attacks on NLP systems generally attempt to search over discrete tokens and, instead of bounding the perturbation in Euclidean space, use linguistic constraints over the selected tokens. 

\citet{alzantot2018generating} implement a discrete genetic algorithm where genes are  tokenized sentences (Genetic Alg in Table \ref{tab:all_results}). Two genes can crossover by sampling a parent for each position in the sentence and selecting the corresponding token. Mutations are performed by sampling from nearby tokens in embedding space. The embedding space they use is the counter-fitted GloVe embeddings \citep{pennington2014glove,mrksic-etal-2016-counter}. \citet{zang-etal-2020-word} also work with a discretized search algorithm, in their case Particle Swarm Optimization (PSO in Table \ref{tab:all_results}). Each particle is again a tokenized sentence and has a velocity associated with each token. Instead of moving in continuous space according to its velocity in each step, a position in the sentence ``jumps'' from one token to another. Instead of operating over embeddings, they use the lexicon HowNet \cite{dong2003hownet} to find synonyms of current words.
\citet{jin2020bert} create a powerful baseline for adversarial attacks which first determines an importance ranking over words and then changes each token in order (Word Importance Ranking, WIR in Table \ref{tab:all_results}). Tokens can be changed to one of the $50$ nearest neighbours in embedding space but are subject to constraints. The replacement token must have a cosine similarity of at least $0.7$ in the attack's embedding space, again in the counter-fitted GloVe embedding space by \citet{mrksic-etal-2016-counter}. Additionally, replacement tokens must have the same part of speech and a sentence level similarity score given by a Universal Sentence Encoder \citep{cer2018universal} above a given threshold. The word importance ranking is determined by the difference in output when a given token is removed from the input.

The above black-box attacks on NLP systems along with greedy and beam search are re-implemented by \citet{morris2020textattack}. Their library, TextAttack, seeks to be a test bed for comparing adversarial attacks on NLP systems. To this end, a companion paper comparing different attacks has been released by \citet{yoo2020searching}. 

\section{Zeroth Order Optimization-Based Attack (ZOO)}

\citet{chen2017zoo} work under the adversarial attack framework of \citet{carlini2017towards} and use a zeroth order optimization method to work in a black-box setting to attack images. The ZOO idea is to create pseudogradients using zeroth order estimates of the true gradient. Their method performs point-wise perturbations on individual pixels by adding a scaled one-hot vector $e_i$ to the image $x$ and computing the difference between the outputs. The derivative of this single perturbed pixel can then be calculated by $\frac{f(x + \mu e_i) - f(x - \mu e_i)}{2\mu}e_i$, where $\mu$ controls how far apart the two queried points are. 

\citet{chen2017zoo} use coordinate descent \citep{poljak1973pseudogradient} and perturb a single pixel at a time to construct a gradient. Instead of having to query the model $2m$ times, this method can be optimized by perturbing all pixels simultaneously with a vector $u$ sampled from a multivariate Gaussian distribution \citep{NesterovSpokoiny:15}. This two-point function evaluation in expectation approximates the true gradient of the function but can also be applied in situations where the function is unknown but smooth, which is the case in black-box attacks on neural networks. The estimated gradient can then be used in standard gradient descent: 
$$
x_{t+1} = x_{t} - \lambda \frac{1}{n} \sum_{i = 1}^n \frac{f(x + \mu u_i) - f(x)}{\mu}u_i.
$$
Inspired by the zeroth order optimization-based attack of \citet{chen2017zoo} and the interpretable displacements of \citet{sato2018interpretable}, we developed a discretized version of zeroth order optimization for attacks on NLP systems, dubbed DiscreteZOO.

\section{DiscreteZOO for Black-Box Attacks on NLP Systems}

Zeroth order optimization assumes nothing about the function to optimize other than the fact that it is smooth. We add the assumptions here that we possess a surrogate embedding space and that there is a smooth transformation from this embedding space to the internal embedding space of the target model. Continuous zeroth order gradient estimation requires that any arbitrary point be queryable in order to use Gaussian noise as perturbations. In the discrete case, however, the system only allows us to query with tokens. Instead of sampling from a Gaussian distribution then, we sample from the nearest neighbours around the target token in the surrogate embedding space. 

Given the current token $t$, we can perturb its embedding $e_t$ with a vector $\mu u_{k,t}$, with $\mu = ||e_k - e_t||$, $u_{k,t} = \frac{e_k - e_t}{\mu}$, and $e_k$ being the embedding of another token. 
We can use this to calculate the finite differences between two different function evaluations. 
Sampling a set of multiple neighbouring tokens $\mathcal{N}$ for displacements yields the update rule \[e_{t+1} = e_t - \lambda \frac{1}{|\mathcal{N}|}\sum_{k \in \mathcal{N}} \frac{f(e_t + \mu u_{k,t}) - f(e_t)}{\mu}u_{k,t}.\]
Adding this estimated direction vector to the current embedding then moves towards an area of increasing goal function values, where we would hope to find a token suitable for flipping the label. Once the direction vector is added to the embedding, we snap it to the nearest existing token by cosine similarity. 

In addition to the ZOO-inspired search, we use the word importance ranking of \citet{jin2020bert} and \citet{yoo2020searching}, which replaces each token in the text with an $\texttt{unk}$ token to decide which tokens to attack first. Additionally, we noticed that often label-flipping tokens are already available within the sampled tokens $\mathcal{N}$. In this case, we should accept the sampled token and return early, saving queries to the model, instead of continuing to sample and construct a direction vector. Another optimization prevents the algorithm from snapping to the nearest token if that token decreases the goal function value. 

A detailed algorithm is given in the appendix. Code implementing the algorithm can be found at \url{https://github.com/StatNLP/discretezoo}.

\section{Experiments}

\subsection{Baselines}

In addition to the methods from \citet{yoo2020searching}, we also implement a \textit{random} baseline. Firstly, instead of determining the importance by deleting each token or replacing it with \texttt{unk}, the indices are randomly shuffled and taken as the attack order.  Secondly, replacement tokens are chosen randomly as well. In the \textit{constr.} setup, the random choice is over the Top-N tokens ranked by proximity in the counter-fitted GloVe embedding space \citep{mrksic-etal-2016-counter}, filtered by the constraints. In the \textit{unconstr.} setup, the same method is used to generate the list of tokens, but no filtering is applied. Instead of iterating over them and selecting the token with the highest goal function value, the random baseline samples a single token from the list, inserts into the sentence, and returns if the token flips the label along with the goal function value. If the goal function values is improved then the token is kept and if not it is discarded--the goal function value is returned with the label and does not change the query count. If the label is not flipped, it moves on to the next target position.

Two additional baselines are picking the \textit{farthest} or \textit{closest} token at each step. They function similarly to the random baseline, but instead of sampling from the list of replacement tokens, they immediately pick the farthest or closest token in embedding space, respectively. If all tokens that pass the constraints are equally acceptable, then picking the most dissimilar acceptable token should be a good heuristic. 

The baselines and the WIR methods \cite{yoo2020searching} all iterate over the indices of the target sentence once. In order to see the effects of random continued sampling until the label flips, we also introduce a method called \textit{random, CS}. This method iterates over the indices multiple times and samples tokens until a label flip is achieved or until a maximum number of tokens has been sampled. For this method to work, the \texttt{RepeatModification} constraint has to be removed. Therefore, it is a stronger attack, as it can make changes that are not available to the other methods. Still, this should illustrate the upper bound achievable with random sampling and keeping tokens that improve the goal function.    

\subsection{Constraints}

In \citet{yoo2020searching}, a number of constraints are used to maintain semantic similarity of the adversary with the original text. The main constraints of interest are cosine similarity, BERTScore \citep{zhang2015character}, and part of speech tagging. In the \textit{constr.} setting, replacement tokens must have a cosine similarity of $0.9$ with the original token, the two sentences must have a BERTScore of $0.9$, and the part of speech can not be changed. We also introduce a \textit{unconstr.} setting, where all constraints are removed.

\subsection{Target Models and Goal Functions}

All of the attacks in this paper target multiclass BERT models \citep{devlin2019bert} fine-tuned on SNLI \citep{bowman-etal-2015-large} or Movie Reviews \citep{10.3115/1219840.1219855}. Both are provided by the TextAttack library \citep{morris2020textattack} and are used in the comparison done by \citet{yoo2020searching}.

\citet{morris2020textattack} use a goal function that seeks to minimize the probability of the true label:
\[\mathcal{G}_{TextAttack} = 1 - f(x)_l,\] 
where $f(x)$ is the neural network with a probability vector as output and $l$ is the index of the true label. Working from the zeroth order optimization-based attack of \citet{chen2017zoo}, we also implement their goal function: 
\[\mathcal{G}_{zoo} = - \max(\log f(x)_{l} - \max_{l' \neq l}\log f(x)_{l'}, -\kappa).\]
In our experiments we found no difference in running all attacks on both goal functions, so here we report only results on the $zoo$ loss.

\tabcolsep=0.1cm
 \begin{table*}[t!]
     \centering
     \begin{tabular}{|c|c|rr|rr|rr|}
     \hline
         $\qquad$  & Method & \multicolumn{2}{c}{Attack Success\% $\uparrow$ ($\pm\sigma$)} & \multicolumn{2}{c}{Avg \# Queries $\downarrow$ ($\pm\sigma$)} & \multicolumn{2}{c}{\% Words Changed $\downarrow$ ($\pm\sigma$)} \\
          &  & constr. & unconstr. & constr. & unconstr. & constr. & unconstr. \\
    \hline
    \hline
          \parbox[t]{1mm}{\multirow{15}{*}{\rotatebox[origin=c]{90}{Movie Reviews}}} & Greedy & 20.6 &  100.0 & 35.0 & 1111.5 & 9.6 & 9.2 \\
              & Beam 4 & 21.4 & 100.0 & 95.1 & 1688.1 & 10.0 & 8.9 \\ 
              & Beam 8 & 21.8 & 100.0 & 174.9 & 2504.1 & 10.1 & 8.9 \\ 
              & WIR(\texttt{UNK}) & 17.8 & 99.6 & 30.1 & 153.8 & 11.8 & 16.6 \\
              & WIR(\texttt{DEL}) & 17.0 & 99.8 & 30.2 & 147.5 & 11.7 & 16.4 \\
              & WIR(\texttt{PWWS}) & 18.6 & 99.6 & 44.2 & 894.4 & 10.6 & 10.0 \\
              & WIR(\texttt{RAND}) & 16.8±0.5 & 99.6 & 12.5±0.1 & 172.8 & 12.9±0.5 & 20.2\\
              & Genetic Alg & 19.4 & 99.6 & 2658.6 & 2578.6 & 10.6 &  10.3 \\
              & PSO & 22.0 & 100.0 & 2456.4 & 4010.2 & 11.1 & 10.5 \\
              & \textbf{DiscrZOO} & 20.4±0.0 & 97.9±0.4 & 33.2±0.0 & 118.7±2.5 & 13.7±0.0 & 17.8±0.2 \\
              & Random & 11.5±0.7 & 68.3±1.8 & 5.4±0.0 & 12.2±0.2 & 11.9±0.7 & 50.6±1.0\\
              & Random, CS & 17.3±0.2 & 98.7±0.6 & 15.1±0.1 & 15.9±0.4 & 13.9±0.4 & 37.5±0.4 \\
              & Farthest  & 12.3±0.7 & 78.3±0.3 & 5.3±0.0 & 10.6±0.1 & 11.2±0.3 & 27.1±0.4\\
              & Closest & 11.1±0.4 & 55.5±0.7 & 5.4±0.0 & 13.1±0.2 & 12.2±0.7 & 30.0±1.0 \\
     \hline
     \hline
          \parbox[t]{1mm}{\multirow{15}{*}{\rotatebox[origin=c]{90}{SNLI}}} & Greedy & 19.9 & 100.0 & 7.2 & 422.0 & 6.1 & 5.9 \\
              & Beam 4 & 20.1 & 100.0 & 12.3 & 563.0 & 6.1 & 5.9 \\ 
              & Beam 8 & 20.1 & 100.0 & 17.8 & 739.0 & 6.1 & 5.8 \\ 
              & WIR(\texttt{UNK}) & 19.6 & 100.0 & 26.2 & 84.9 & 7.0 & 6.5 \\
              & WIR(\texttt{DEL}) & 18.7 & 100.0 & 26.2 & 85.3 & 7.3 & 6.6 \\
              & WIR(\texttt{PWWS}) & 19.1 & 99.9 & 30.5 & 407.2 & 6.0 & 6.8 \\
              & WIR(\texttt{RAND}) & 18.3±0.1 & 100.0 & 5.0±0.0 & 85.9 & 7.3±0.2 & 8.7\\
              & Genetic Alg & 19.7 & 99.9 & 1328.5 & 998.5 & 6.1 & 6.4 \\
              & PSO & 20.0 & 100.0 & 1262.4 & 815.2 & 6.2 & 6.4 \\
              & \textbf{DiscrZOO} & 21.2±0.1 & 98.1±0.2 & 26.9±0.0 & 58.3±0.1 & 6.7±0.0 & 6.8±0.1 \\
              & Random  & 10.1±0.3 & 70.0±0.4 & 2.6±0.0 & 5.6±0.0 & 7.3±0.2 & 17.7±0.4 \\
              & Random, CS & 18.7±0.4 & 99.9±0.1 & 6.8±0.04 & 7.3±1.0 & 8.0±0.1 & 16.3±0.2\\
              & Farthest & 11.6±0.2 & 81.0±0.7 & 2.6±0.0 & 5.1±0.0 & 7.2±0.2 & 12.3±0.2\\
              & Closest & 7.9±0.2 & 58.0±0.7 & 2.7±0.0 & 6.4±0.1 & 7.4±0.3 & 15.8±0.1 \\
     \hline
     \end{tabular}
     \caption{Adversarial attack using ZOO loss on both Movie Reviews and SNLI. DiscreteZOO averaged over three runs; Random, Farthest, and Closest Nearest Neighbor baselines averaged over seven runs; standard deviation denoted by $\pm$. With constraints, the DiscreteZOO attack samples up to 10 neighbors, while the unconstrained attack samples up to 25 neighbors.}
     \label{tab:all_results}
 \end{table*}

\section{Results}

Results for the baseline attacks and \textit{DiscreteZOO} are summarized in Table \ref{tab:all_results}. 

In the \textit{constr.} setup on SNLI, \textit{DiscreteZOO} outperforms all other methods in terms of success rate. On Movie Reviews,  \textit{DiscreteZOO} achieves slightly higher results than the \textit{WIR} methods with comparable query counts, but it is not as successful as the \textit{beam} search or \textit{PSO} methods.
For \textit{DiscreteZOO}, the standard deviation is either very low or actually zero. This suggests that there is not a large enough space from which the algorithm can sample. The \textit{farthest} baseline already achieves a success rate of $12.3\%$ on Movie Reviews and $11.6\%$ on SNLI, while \textit{random} and \textit{closest} achieve 11.5\% and 11.1\% on Movie Reviews and 10.1\% and 7.9\% on SNLI, respectively. Thus, simple heuristics already accomplish a substantial amount of the achievable success rates at minimal query count, while the generally low success rates across all algorithms suggests that the constraints are too strict, preventing exploitation of more sophisticated search techniques. The \textit{continued sampling} version of the random baseline achieves a success rate of $17.3\%$ on Movie Reviews and $18.7\%$ on SNLI. This is directly competitive with the search based methods.

In the unconstrained setup, all search methods, including \textit{DiscreteZOO}, approach success rates of $100\%$. The methods are only able to evaluate the $50$ nearest neighbors, however, the \textit{farthest} baseline reaches success rates of 78.31\% and 81.04\% on Movie Reviews and SNLI, respectively. \textit{Random} and \textit{closest} follow \textit{farthest} in terms of success rate, with a query count that is an order of magnitude lower than other methods. The \textit{continued sampling} version of the random baseline achieves nearly $100\%$ success rate on both datasets with a similarly small query count. 
The relative success of the baseline heuristic methods shows that the nearest neighbor structure in the embedding space is already powerful enough to flip the label. Much of the success of the other methods can already be attributed to these simple heuristics, showing that search is not always necessary.

\section{Analysis}

The results of the previous experiments show slight improvement over previous methods but they appear curious. For example, some of the results have no variation on repeated runs even with a stochastic algorithm. Additionally, random token selection appears to be competitive with \textit{greedyWIR} methods.

The counterfitted GLoVE embedding space \citep{mrksic-etal-2016-counter} is  a very sparse embedding space, containing a total of $65713$ tokens. Using the neighborhood threshold given by the \textit{constr.} constraints, a cosine similarity of $0.9$ or higher, then the average token in this embedding space has $0.72$ neighbors. Among tokens that have neighbors, the average rises to $2.63$. Histograms showing number of tokens over neighborhood size with different neighborhood definitions can be seen in figure \ref{fig:neighborhood_graphs} in the appendix. These values are an upper bound on the actual average number of neighbors as the constraints also include BERTScore and a Part of Speech constraint which further restrict the space. 

The sparsity of the space coupled with the size of the neighborhoods induced by cosine similarity of $0.9$ or higher could explain why the naive methods are able to perform so well compared to more sophisticated optimization methods. There are so few valid replacement tokens in the space that it is entirely feasible to try every option with the \textit{greedyWIR} or \textit{beam} methods. 

With a large enough sample size, \textit{discretezoo} already observes all allowed replacement tokens during sampling and is able to stop if one of them flips the label. In this case, \textit{discretezoo} and \textit{greedyWIR} should produce similar results. This is effectively demonstrated by the \textit{random} baseline, which chooses a replacement token randomly and is unable to use information from the goal function to decide between multiple replacement tokens. Because there are so few possible replacement tokens, randomly selecting from the few that are available is roughly as good of a strategy as beam search or greedily selecting the best replacement token from all options. The results given by the random baseline are close to those reported by \citet{yoo2020searching} for their \textit{greedyWIR} baselines. This suggests that the success of their attacks compared to the more advanced search methods is not because of their algorithm finding good texts but rather because the search space has been so restricted that any choice will work. Methods besides \textit{greedyWIR} are allowed more flexibility. Instead of just considering one target position at a time, they are allowed to consider the best replacement for all positions. This turns the attack into a combinatorial problem of finding which combination of positions is best to attack instead of which replacement tokens are the best.

Additionally, this shows that the models being attacked are very brittle. Given a single target token that has replacements, there are on average only three to choose from. Selecting one of these three very similar tokens is already enough to flip the label in many cases. 

\section{Conclusion}

Zeroth order optimization methods have been shown to yield superior performance for black box attacks in continuous spaces such as images \cite{chen2017zoo}. 
In attacks on NLP systems, only a discrete set of tokens are admissable as input but they are still processed as vectors in continuous space. This allows for optimization in the continuous space instead of in the discrete token space.
We implement a zeroth order optimization algorithm in the TextAttack library \citep{morris2020textattack} and compare with the results of \citet{yoo2020searching}. While our method appears to be competitive, we find that the linguistic constraints imposed on the search methods are so tight that nearly no optimization is necessary.  Instead, selecting the farthest allowable token or a random token is already enough to flip the label in many cases. We argue that more robust tasks are required for meaningful research on black-box adversarial text attacks.

\bibliographystyle{acl_natbib}
\bibliography{custom}

\appendix

\clearpage

\section{Appendix}

\subsection{Algorithms}

\begin{algorithm}[h!]
    \footnotesize
    \DontPrintSemicolon
    \SetAlgoLined    
    \KwResult{Returns a pair $(successful, \hat{s})$, containing a boolean indicating success and a sequence of tokens}
    \KwIn{$s$ is the original sentence to attack, $T = [ t_0,  t_1, ..., t_n ]$ is an ordered sequence of indices to target, filtered by pre-transformation constraints, $n$ is the number of replacement tokens to use for calculating displacements, and $u$ is the number of gradient updates the algorithm can perform.}
    \KwData{$E$ is a matrix of word embeddings, $\texttt{goal\_function}$ is a function that returns goal function value and model prediction.}
     $y = \text{argmax}\,model(s)$\;
     $\hat{s} = s$\;
     \For{$t \; \text{in} \; T$}{
        $steps = 0$\;
         \While{$steps < u$}{
        $\overrightarrow{e_t}$ = $E_{s_t}$\;
        $\text{neighborhood} = \texttt{knn}(\overrightarrow{e_t}, E, k \cdot n)$\;
        $\text{candidates} = \texttt{choose}(\text{neighborhood}, n)$\;
        $(l, \hat{y}) = \texttt{goal\_function}(\hat{s}, y)$\;
        $\overrightarrow{sumd} = \overrightarrow{0}$\;
        \For{$c \in \text{candidates}$}{
            $\hat{s}_t^c = \{s_0,..., s_{t-1}, c, s_{t+1},...\}$\;
            $(l_{c}, y_{c}) = \texttt{goal\_function}(\hat{s}_t^c, y)$\;
            \If{$y_c \neq y$}{
                \Return $(\texttt{True}, \hat{s}_t^c)$
            }
            $\overrightarrow{sumd} = \overrightarrow{sumd} + \frac{l_c - l}{\norm{\overrightarrow{e_c} - \overrightarrow{e_t}}}\frac{\overrightarrow{e_c} - \overrightarrow{e_t}}{\norm{\overrightarrow{e_c} - \overrightarrow{e_t}}}$\;
        }
        $\text{r} = \texttt{knn}(e_t + \gamma \cdot \overrightarrow{sumd}, E, 1)$\;
        $\Tilde{s} = \{\hat{s}_0,...\hat{s}_{t-1}, r, \hat{s}_{t+1},...\}$\;
        $(\Tilde{l}, \Tilde{y}) = \texttt{goal\_function}(\Tilde{s}, y)$\;
        \If{$\Tilde{l} \geq l$}{
            $\hat{s} = \Tilde{s}$\;
            $\hat{y} = \Tilde{y}$\;
        }
        \If{$\hat{y} \neq y$}{
                \Return $(\texttt{True}, \hat{s})$
            }
        $steps = steps + 1$; 
        }
     }
    \Return $(\texttt{False}, \hat{s})$
     \caption{Unconstrained Attack}
     \label{algorithm:unconstrained_attack}
\end{algorithm}

\begin{algorithm}[h!]
    \footnotesize
    \DontPrintSemicolon
    \SetAlgoLined    
    \KwResult{Returns a pair $(successful, \hat{s})$, containing a boolean indicating success and a sequence of tokens}
    \KwIn{$s$ is the original sentence to attack, $T = [ t_0,  t_1, ..., t_{n-1} ]$ is an ordered sequence of indices to target, filtered by pre-transformation constraints. $b$ is a maximum number of queries to perform.}
    \KwData{$E$ is a matrix of word embeddings, $\texttt{goal\_function}$ is a function that returns goal function value and model prediction.}
     $y = \text{argmax}\,model(s)$\;
     $\hat{y} = y$\;
     $\hat{s} = s$\;
     \For{$t$ in $T$}{
        $\text{current\_token} = \hat{s}_t$\;
        $\text{current\_embedding} = E_{\text{current\_token}}$\;
        $\text{replacements} = \texttt{knn}(\text{current\_embedding}, E, 50)$\;
        $r = \texttt{shuffle}(replacements)_0$\;
        $\Tilde{s} = \{\hat{s}_0,...\hat{s}_{t-1}, r, \hat{s}_{t+1},...\}$\;
        $(\Tilde{l}, \Tilde{y}) = \texttt{goal\_function}(\Tilde{s}, y)$\;
        \If{$\Tilde{l} \geq l$}{
            $\hat{s} = \Tilde{s}$\;
            $\hat{y} = \Tilde{y}$\;
        }
        \If{$\hat{y} \neq y$}{
                \Return $(\texttt{True}, \hat{s})$
            }
     }
    \Return $(\texttt{False}, \hat{s})$
     \caption{Random Baseline}
     \label{algorithm:random_baseline}
\end{algorithm}

\begin{algorithm}[h!]
    \footnotesize
    \DontPrintSemicolon
    \SetAlgoLined    
    \KwResult{$(successful, \hat{s})$}
    \KwIn{$s$, $T = [ t_0,  t_1, ..., t_{n-1} ]$, $b$}
    \KwData{$E$, $\texttt{goal\_function}$}
     $y = \text{argmax}\,model(s)$,
     $\hat{y} = y$,
     $\hat{s} = s$,
     $count = 0$,
     $ci = 0$\;
     \While{$y = \hat{y} \wedge count < b$}{
        \If{$ci \geq n $}{
            $ci = 0$\;
        }
        $\text{current\_target} = t_{ci}$\;
        $\text{current\_token} = \hat{s}_{\text{current\_target}}$\;
        $\text{current\_embedding} = E_{\text{current\_token}}$\;
        $\text{replacements} = \texttt{knn}(\text{current\_embedding}, E, 50)$\;
        $r = \texttt{shuffle}(replacements)_0$\;
        $\Tilde{s} = \{\hat{s}_0,...\hat{s}_{t_{ci}-1}, r, \hat{s}_{t_{ci}+1},...\}$\;
        $(\Tilde{l}, \Tilde{y}) = \texttt{goal\_function}(\Tilde{s}, y)$\;
        $count = count + 1$\;
        \If{$\Tilde{l} \geq l$}{
            $\hat{s} = \Tilde{s}$\;
            $\hat{y} = \Tilde{y}$\;
        }
        \If{$\hat{y} \neq y$}{
                \Return $(\texttt{True}, \hat{s})$
            }
        $ci = ci + 1$; 
     }
    \Return $(\texttt{False}, \hat{s})$
     \caption{Random Baseline, Continued Sampling}
     \label{algorithm:random_baseline_cs}
\end{algorithm}

\newpage


\subsection{Embedding Space Analysis}

\begin{figure}[h!]
    \centering
    \includegraphics[width=0.85\textwidth]{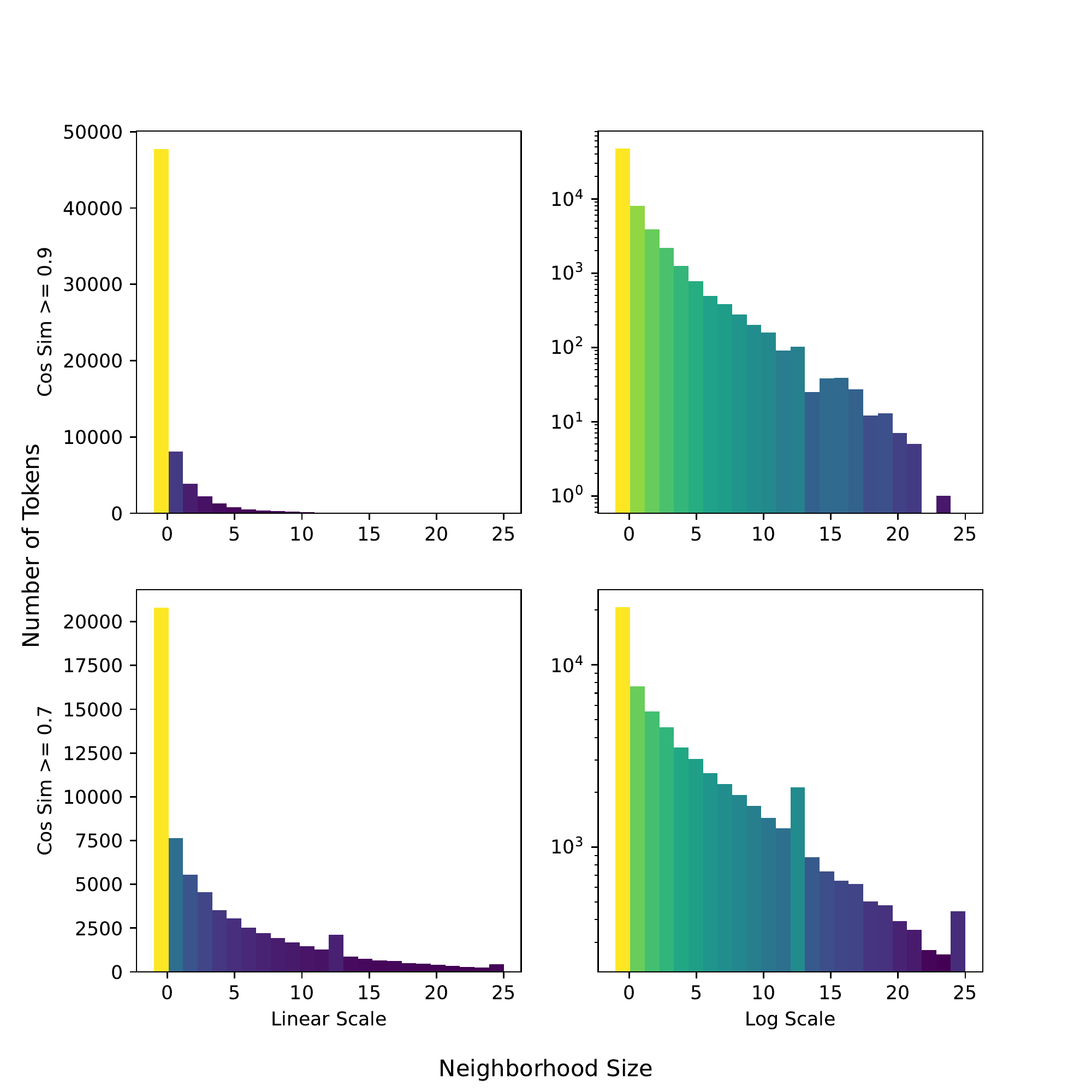}
    \caption{Graphs of number of tokens over number of neighbors, under two neighborhood definitions. Both linear and log scales are used in order to highlight outlying values.}
    \label{fig:neighborhood_graphs}
\end{figure}

\subsection{Relaxed Constraint Results}

\include{lax_results_table_new}

\end{document}

%% file: lax_results_table_new.tex
\tabcolsep=0.1cm
 \begin{table*}[h!]
     \centering
     \begin{tabular}{|c|c|r|r|r|}
     \hline
         $\qquad$  & Method & \multicolumn{1}{c}{Attack Success\% $\uparrow$ ($\pm\sigma$)} & \multicolumn{1}{c}{Avg \# Queries $\downarrow$ ($\pm\sigma$)} & \multicolumn{1}{c}{\% Words Changed $\downarrow$ ($\pm\sigma$)} \\
          & & lax. & lax. & lax.\\
    \hline
    \hline
          \parbox[t]{1mm}{\multirow{13}{*}{\rotatebox[origin=c]{90}{Movie Reviews}}} & Greedy & 84.6 & 208.93 & 14.2 \\
              & Beam 4 & 89.4 & 525.3 & 14.5 \\ 
              & Beam 8 & 90.2 & 933.3 & 14.3 \\ 
              & WIR(\texttt{UNK}) & 78.2 & 67.3 & 19.8 \\
              & WIR(\texttt{DEL}) & 78.8 & 67.5 & 20.1 \\
              & WIR(\texttt{PWWS}) & 75.6 & 142.1 & 17.6 \\
              & WIR(\texttt{RAND}) & 76.7±0.8 & 55.0±0.6 & 22.9±0.5 \\
              & \textbf{DiscrZOO} & 60.5±0.5 & 66±0.3 & 17.0±0.2  \\
              & Random & 33.6±0.6 & 8.5±0.0 & 20.4±0.4 \\
              & Random, CS & 63.9±1.2 & 17.7±0.3 & 27.9±0.3 \\
              & Farthest & 36.1±0.8 & 8.4±0.6 & 20.6±0.6 \\
              & Closest & 27.0±0.6 & 8.8±0.0 & 20.1±0.2 \\
     \hline
     \hline
          \parbox[t]{1mm}{\multirow{13}{*}{\rotatebox[origin=c]{90}{SNLI}}} & Greedy & 89.1 & 49.24 & 6.6 \\
              & Beam 4 & 90.0 & 78.37 & 6.7 \\ 
              & Beam 8 & 90.1 & 117.0 & 6.7 \\ 
              & WIR(\texttt{UNK}) & 88.1 & 39.8 & 8.1 \\
              & WIR(\texttt{DEL}) & 88.4 & 39.9 & 8.4 \\
              & WIR(\texttt{PWWS}) & 86.8 & 65.6 & 7.0 \\
              & WIR(\texttt{RAND}) & 87.3±0.4 & 21.9±0.2 & 9.6±0.1 \\
              & \textbf{DiscrZOO} & 81.7±0.3 & 37.1±0.1 & 7.0±0.1 \\
              & Random  & 47.3±1.1 & 4.3±0.0 & 10.0±0.2 \\
              & Random, CS & 84.5±0.4 & 9.5±0.1 & 12.7±0.3 \\
              & Farthest & 57.2±0.5 & 4.1±0.0 & 10.0±0.2 \\
              & Closest & 32.0±0.4 & 4.6±0.0 & 10.2±0.1 \\
     \hline
     \end{tabular}
     \caption{Adversarial attack using ZOO loss on both Movie Reviews and SNLI. DiscreteZOO averaged over three runs; Random, Farthest, and Closest Nearest Neighbor baselines averaged over seven runs; standard deviation denoted by $\pm$. With constraints, the DiscreteZOO attack samples up to 10 neighbors. The \textit{lax.} constraints are the same as the \textit{constr.} from Table \ref{tab:all_results} with the cosine similarity and BERTScore thresholds set to $0.7$.}
     \label{tab:lax_results}
 \end{table*}

%% file: emnlp2021.bbl
\begin{thebibliography}{19}
\expandafter\ifx\csname natexlab\endcsname\relax\def\natexlab#1{#1}\fi

\bibitem[{Alzantot et~al.(2018)Alzantot, Sharma, Elgohary, Ho, Srivastava, and
  Chang}]{alzantot2018generating}
Moustafa Alzantot, Yash Sharma, Ahmed Elgohary, Bo-Jhang Ho, Mani Srivastava,
  and Kai-Wei Chang. 2018.
\newblock Generating natural language adversarial examples.
\newblock In \emph{EMNLP}.

\bibitem[{Bowman et~al.(2015)Bowman, Angeli, Potts, and
  Manning}]{bowman-etal-2015-large}
Samuel~R. Bowman, Gabor Angeli, Christopher Potts, and Christopher~D. Manning.
  2015.
\newblock A large annotated corpus for learning natural language inference.
\newblock In \emph{EMNLP}.

\bibitem[{Carlini and Wagner(2017)}]{carlini2017towards}
Nicholas Carlini and David Wagner. 2017.
\newblock Towards evaluating the robustness of neural networks.
\newblock In \emph{IEEE S\&P}.

\bibitem[{Cer et~al.(2018)Cer, Yang, Kong, Hua, Limtiaco, John, Constant,
  Guajardo-Cespedes, Yuan, Tar et~al.}]{cer2018universal}
Daniel Cer, Yinfei Yang, Sheng-yi Kong, Nan Hua, Nicole Limtiaco, Rhomni~St
  John, Noah Constant, Mario Guajardo-Cespedes, Steve Yuan, Chris Tar, et~al.
  2018.
\newblock Universal sentence encoder for english.
\newblock In \emph{EMNLP Demo}.

\bibitem[{Chen et~al.(2017)Chen, Zhang, Sharma, Yi, and Hsieh}]{chen2017zoo}
Pin-Yu Chen, Huan Zhang, Yash Sharma, Jinfeng Yi, and Cho-Jui Hsieh. 2017.
\newblock {ZOO}: Zeroth order optimization based black-box attacks to deep
  neural networks without training substitute models.
\newblock In \emph{Workshop on Artificial Intelligence and Security}.

\bibitem[{Devlin et~al.(2019)Devlin, Chang, Lee, and
  Toutanova}]{devlin2019bert}
Jacob Devlin, Ming-Wei Chang, Kenton Lee, and Kristina Toutanova. 2019.
\newblock Bert: Pre-training of deep bidirectional transformers for language
  understanding.
\newblock In \emph{NAACL}.

\bibitem[{Dong and Dong(2003)}]{dong2003hownet}
Zhendong Dong and Qiang Dong. 2003.
\newblock Hownet-a hybrid language and knowledge resource.
\newblock In \emph{International Conference on Natural Language Processing and
  Knowledge Engineering}, pages 820--824, Beijing, China.

\bibitem[{Jin et~al.(2020)Jin, Jin, Zhou, and Szolovits}]{jin2020bert}
Di~Jin, Zhijing Jin, Joey~Tianyi Zhou, and Peter Szolovits. 2020.
\newblock Is bert really robust? a strong baseline for natural language attack
  on text classification and entailment.
\newblock In \emph{AAAI}.

\bibitem[{Morris et~al.(2020)Morris, Lifland, Yoo, Grigsby, Jin, and
  Qi}]{morris2020textattack}
John Morris, Eli Lifland, Jin~Yong Yoo, Jake Grigsby, Di~Jin, and Yanjun Qi.
  2020.
\newblock {TextAttack}: A framework for adversarial attacks, data augmentation,
  and adversarial training in {NLP}.
\newblock In \emph{EMNLP Demo}.

\bibitem[{Mrk{\v{s}}i{\'c} et~al.(2016)Mrk{\v{s}}i{\'c}, {\'O}~S{\'e}aghdha,
  Thomson, Ga{\v{s}}i{\'c}, Rojas-Barahona, Su, Vandyke, Wen, and
  Young}]{mrksic-etal-2016-counter}
Nikola Mrk{\v{s}}i{\'c}, Diarmuid {\'O}~S{\'e}aghdha, Blaise Thomson, Milica
  Ga{\v{s}}i{\'c}, Lina~M. Rojas-Barahona, Pei-Hao Su, David Vandyke,
  Tsung-Hsien Wen, and Steve Young. 2016.
\newblock Counter-fitting word vectors to linguistic constraints.
\newblock In \emph{NAACL}.

\bibitem[{Nesterov and Spokoiny(2015)}]{NesterovSpokoiny:15}
Yurii Nesterov and Vladimir Spokoiny. 2015.
\newblock Random gradient-free minimization of convex functions.
\newblock \emph{Foundations of Computational Mathematics}.

\bibitem[{Pang and Lee(2005)}]{10.3115/1219840.1219855}
Bo~Pang and Lillian Lee. 2005.
\newblock Seeing stars: Exploiting class relationships for sentiment
  categorization with respect to rating scales.
\newblock In \emph{ACL}.

\bibitem[{Pennington et~al.(2014)Pennington, Socher, and
  Manning}]{pennington2014glove}
Jeffrey Pennington, Richard Socher, and Christopher~D. Manning. 2014.
\newblock \href {http://www.aclweb.org/anthology/D14-1162} {Glove: Global
  vectors for word representation}.
\newblock In \emph{EMNLP}.

\bibitem[{Poljak and Tsypkin(1973)}]{poljak1973pseudogradient}
Boris Poljak and Yakov Tsypkin. 1973.
\newblock Pseudogradient adaptation and training algorithms.
\newblock \emph{Automation and Remote Control}, 34:45--67.

\bibitem[{Sato et~al.(2018)Sato, Suzuki, Shindo, and
  Matsumoto}]{sato2018interpretable}
Motoki Sato, Jun Suzuki, Hiroyuki Shindo, and Yuji Matsumoto. 2018.
\newblock Interpretable adversarial perturbation in input embedding space for
  text.
\newblock In \emph{IJCAI}.

\bibitem[{Szegedy et~al.(2014)Szegedy, Zaremba, Sutskever, Bruna, Erhan,
  Goodfellow, and Fergus}]{szegedy2014intriguing}
Christian Szegedy, Wojciech Zaremba, Ilya Sutskever, Joan Bruna, Dumitru Erhan,
  Ian Goodfellow, and Rob Fergus. 2014.
\newblock Intriguing properties of neural networks.
\newblock In \emph{ICLR}.

\bibitem[{Yoo et~al.(2020)Yoo, Morris, Lifland, and Qi}]{yoo2020searching}
Jin~Yong Yoo, John Morris, Eli Lifland, and Yanjun Qi. 2020.
\newblock Searching for a search method: Benchmarking search algorithms for
  generating nlp adversarial examples.
\newblock In \emph{BlackboxNLP}.

\bibitem[{Zang et~al.(2020)Zang, Qi, Yang, Liu, Zhang, Liu, and
  Sun}]{zang-etal-2020-word}
Yuan Zang, Fanchao Qi, Chenghao Yang, Zhiyuan Liu, Meng Zhang, Qun Liu, and
  Maosong Sun. 2020.
\newblock \href {https://www.aclweb.org/anthology/2020.acl-main.540}
  {Word-level textual adversarial attacking as combinatorial optimization}.
\newblock In \emph{ACL}.

\bibitem[{Zhang et~al.(2015)Zhang, Zhao, and LeCun}]{zhang2015character}
Xiang Zhang, Junbo Zhao, and Yann LeCun. 2015.
\newblock Character-level convolutional networks for text classification.
\newblock In \emph{NIPS}.

\end{thebibliography}
